\documentclass{svproc}
\usepackage{cite} 
\usepackage{url}

\usepackage[utf8]{inputenc} 
\usepackage[T1]{fontenc}    
\usepackage{hyperref}       
\usepackage{booktabs}       
\usepackage{amsfonts}       
\usepackage{nicefrac}       
\usepackage{microtype}      
\usepackage{xcolor}         

\usepackage{graphicx}       

\usepackage{caption}
\usepackage{subcaption}
\usepackage{fancyvrb}      
\usepackage{xcolor}
\usepackage{fancyvrb}
\usepackage{etoolbox}       
\makeatletter
\patchcmd{\@verbatim}
  {\verbatim@font}
  {\verbatim@font\small}
  {}{}
\makeatother

\graphicspath{ {figures/} } 

\begin{document}
\mainmatter              
\title{Neural-guided, Bidirectional Program Search for Abstraction and Reasoning}
\titlerunning{Neural-guided, Bidirectional Program Search}  


%

\author{Simon Alford\inst{1} \and Anshula Gandhi\inst{1}
\and Akshay Rangamani\inst{1} \and Andrzej Banburski\inst{1} \and Tony Wang\inst{1} \and Sylee Dandekar\inst{2} \and John Chin\inst{1} \and Tomaso Poggio\inst{1} \and Peter Chin\inst{3}}
\authorrunning{Simon Alford et al.} 
%
\tocauthor{Simon Alford, Anshula Gandhi, Akshay Rangamani, Andrzej Banburski, Tony Wang, Sylee Dandekar, John Chin, Tomaso Poggio, and Peter Chin}
\institute{Massachusetts Institute of Technology, Cambridge MA 02139, USA,\\
\and
Raytheon BBN Technologies, Cambridge MA 02138
\and Boston University, Boston MA 02215}
\maketitle              

\begin{abstract}
One of the challenges facing artificial intelligence research today is designing systems capable of utilizing systematic reasoning to generalize to new tasks. The Abstraction and Reasoning Corpus (ARC) measures such a capability through a set of visual reasoning tasks.  In this paper we report incremental progress on ARC and lay the foundations for two approaches to abstraction and reasoning not based in brute-force search. We first apply an existing program synthesis system called DreamCoder to create symbolic abstractions out of tasks solved so far, and show how it enables solving of progressively more challenging ARC tasks. Second, we design a reasoning algorithm motivated by the way humans approach ARC. Our algorithm constructs a search graph and reasons over this graph structure to discover task solutions. More specifically, we extend existing execution-guided program synthesis approaches with deductive reasoning based on function inverse semantics to enable a neural-guided bidirectional search algorithm. We demonstrate the effectiveness of the algorithm on three domains: ARC, 24-Game tasks, and a `double-and-add’ arithmetic puzzle. 
\keywords{abstraction, reasoning, program synthesis, neural networks}
\end{abstract}

\section{Introduction}
\makeatletter{\renewcommand*{\@makefnmark}{} \footnotetext{Published as a conference paper at Complex Networks 2021}\makeatother}
The growth and tremendous success of deep learning has catapulted us past
many benchmarks of artificial intelligence. Reaching human and superhuman performance in object recognition, language generation
and translation, and complex games such as Go and Starcraft has pushed the
boundaries of what humans can do and machines cannot \cite{krizhevsky2012imagenet, he2015deep, devlin2019bert, mnih2015human,silver2017mastering,vinyals2019grandmaster}. To continue to make
progress, we must identify and work towards reducing the gaps between human and machine intelligence.

The Abstraction and Reasoning Corpus (ARC), introduced by Fran\c{c}ois Chollet in 2019, captures an important aspect of human intelligence that our current systems are unable to do: 
the ability to systematically and flexibly generalize to new domains \cite{chollet}. 
Chollet argues that
intelligence must be measured not as skill in a particular task, but as
\textit{skill-acquisition efficiency}. General intelligent systems must also
have \textit{developer-aware generalization}, i.e. be able to solve
problems the developer of the system has not encountered before or anticipated.

ARC consists of training, evaluation, and private test sets of 400, 400, and 200 tasks. Each task consists of 2--4 training
examples and one or more test examples. Each training example is an input/output pair of
grids. To solve a task, an agent must determine the relationship between input
and output grids in the training examples, and use this to produce the correct
output grid for each of the test examples, for which the agent is only given the
input grid. Each task is thus a few-shot learning problem, for which the solution is symbolic and rule-based.

\begin{figure}
    \begin{centering}
        \includegraphics[width=0.5\linewidth]{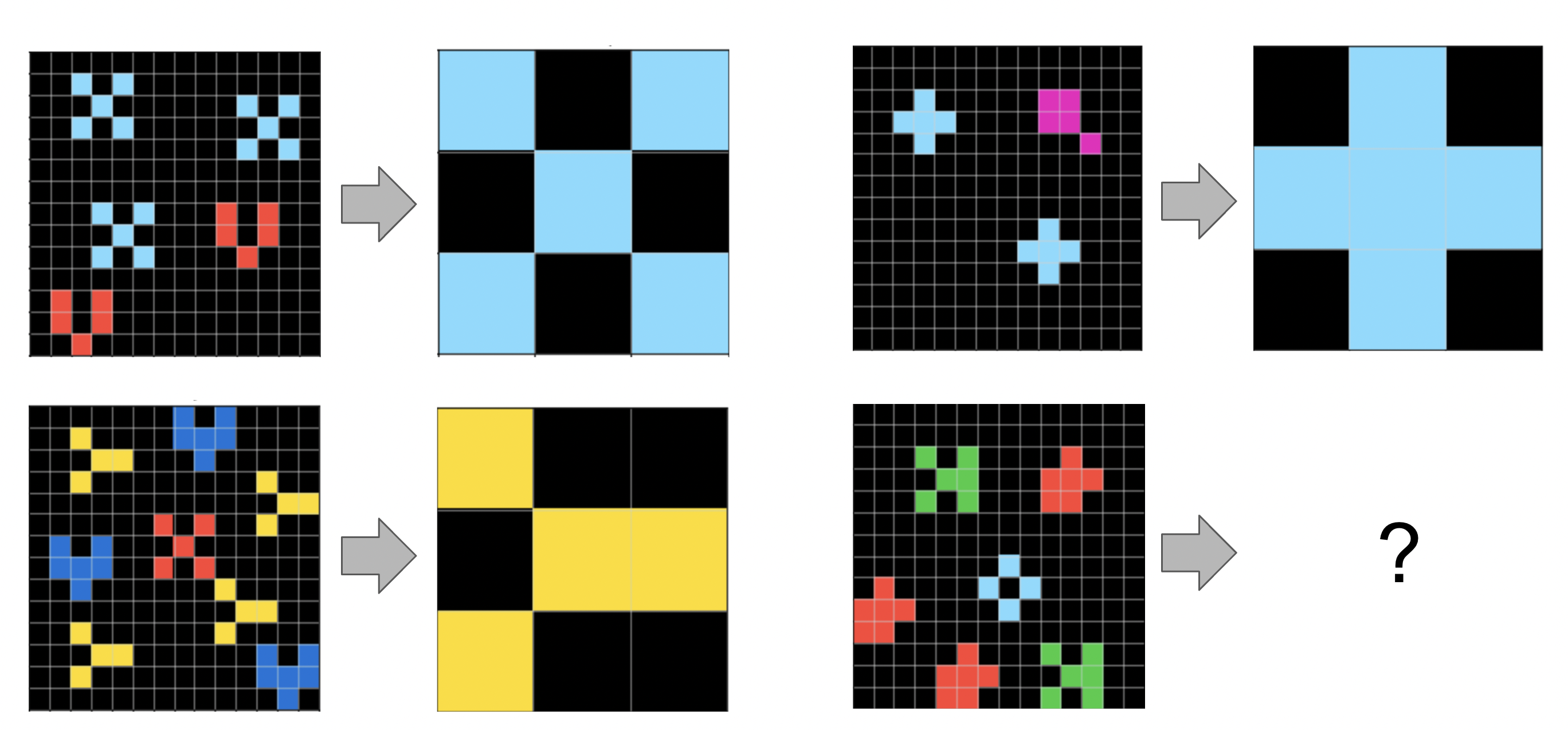}
        \caption[ARC example task]{An example ARC task with three training examples and one test
        example. The solution might be described as ``find the most common
        object in the input grid''.}
        \label{fig:arc_example}
    \end{centering}
\end{figure}

The tasks are unique and constructed by hand so as to prevent
the reverse engineering of any synthetic generation process. They are designed to
depend on a set of human Core Knowledge inbuilt priors such as
objectness, simple arithmetic abilities, symmetry, and goal-directedness.
The evaluation and private test sets are designed such that a solution tailored to the training 
set is unlikely to transfer to the evaluation or test sets. Chollet hosted a Kaggle-competition for ARC and the winning solution, a
hard-coded brute force approach, achieved only $\sim20\%$ performance on the 
private test set \cite{icecuber}.


In this paper we report incremental progress on ARC and lay the foundation for several approaches to abstraction and reasoning not based in brute-force search.
We approach ARC as a program synthesis benchmark,
solving tasks by writing programs that convert input grids to output grids.
In Section \ref{sec:abstraction} we outline an approach to abstraction by applying DreamCoder \cite{ellis2020dreamcoder}.
We show this approach enables learning new concepts that aid in generalization as well as the solving of progressively more challenging tasks.
In Section \ref{sec:reasoning} we describe a novel program synthesis approach motivated by the way humans approach ARC that captures the reasoning required to search for ARC task solutions. Our algorithm constructs a search graph and reasons over this graph structure to discover task solutions. More specifically, we extend existing execution-guided program synthesis approaches \cite{repl, cmu-repl} with deductive reasoning based on function inverse semantics \cite{flashmeta} to enable a neural-guided bidirectional search algorithm. We evaluate our approach on three domains: ARC tasks, `24 Game' problems, and a simple `double-and-add' challenge. These experiments show the benefits of bidirectional search over baselines and the potential for further progress on ARC.
In Section \ref{sec:discussion} we discuss related work, progress on ARC, and future directions.

\section{Abstraction using DreamCoder} \label{sec:abstraction}
We frame the problem as a search problem
over the space of programs expressible in some domain specific language (DSL).
One way a learning agent can achieve {\em developer aware generalization}
(in the sense of \cite{chollet})
is to identify frequently occurring patterns of computation
and form abstractions from them.
These abstractions enable searching for more complex programs more quickly.

In this section we use DreamCoder \cite{ellis2020dreamcoder},
a recent tool for program synthesis, 
to form abstractions.
We first show how DreamCoder's compression algorithm enables learning generalizations of concepts seen in training. Second, we run DreamCoder on ARC to show how forming new abstractions enables 
the agent to solve progressively more challenging tasks.

\subsection{Warmup: Forming Abstractions} \label{subsec:warmup}
\begin{figure}
\centering
\begin{subfigure}{.5\textwidth}
  \centering
  \includegraphics[width=.3\linewidth]{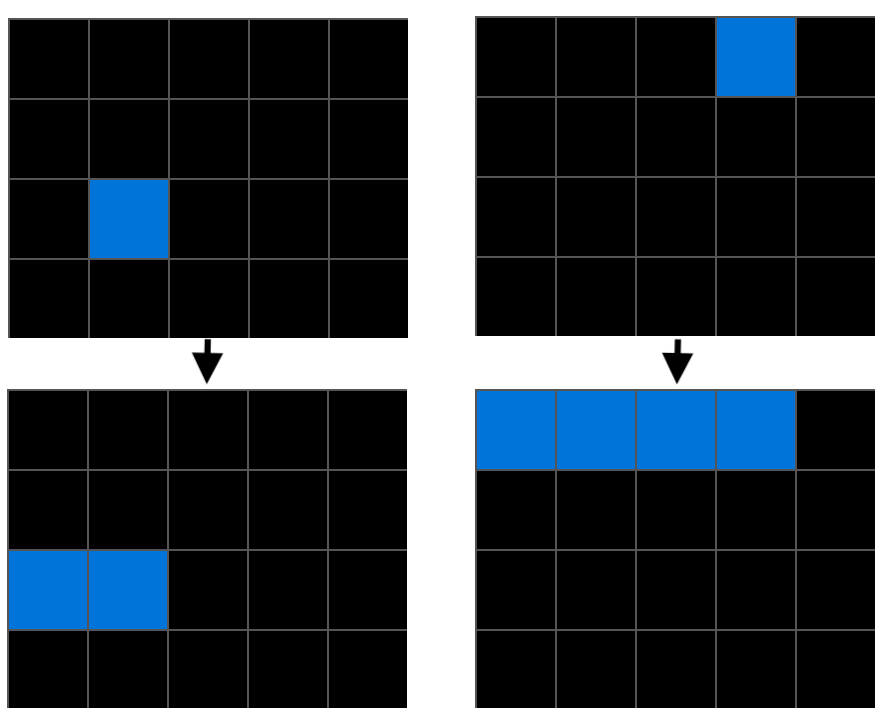}
    \caption[Draw line left task]{An example ``draw line left'' task}
  \label{fig:drawlineleft}
\end{subfigure}%
\begin{subfigure}{.5\textwidth}
  \centering
  \includegraphics[width=.3\linewidth]{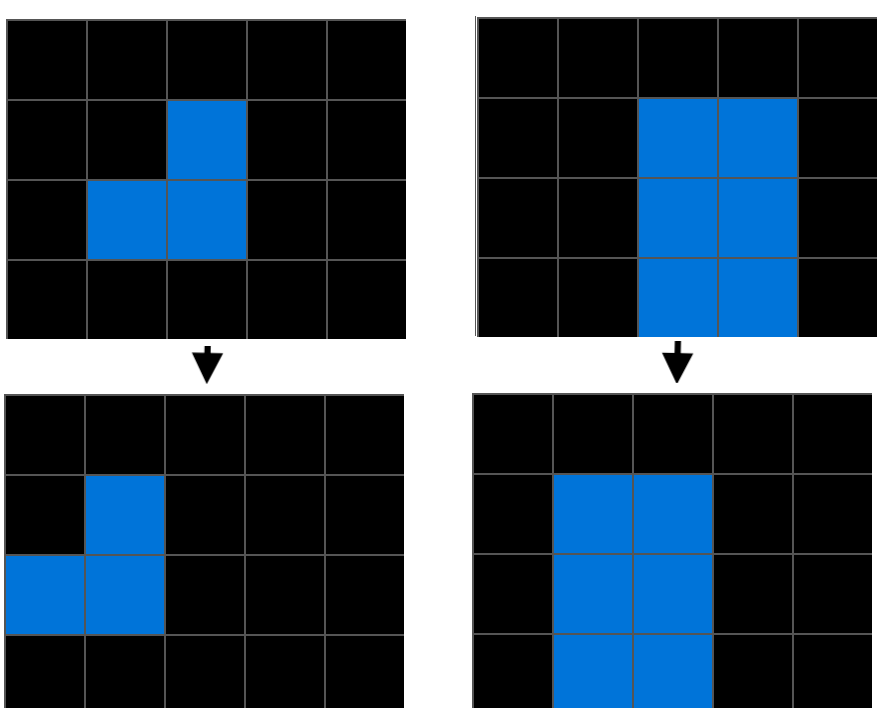}
    \caption[Move object left task]{An example ``move object left'' task}
  \label{fig:moveobjetleft}
\end{subfigure}
\caption{Sample tasks involving applying an action left.}
\label{fig:warmup-sample-tasks}
\end{figure}

To show how DreamCoder can form more abstract concepts from existing ones, we supply our agent with six
synthetic tasks (meant to be similar to ARC tasks): drawing a line in three
different directions, and moving an object in three different directions.
See Figure \ref{fig:warmup-sample-tasks} for a visualization of these tasks.

We solve these tasks with four primitives:
rotate clockwise and counterclockwise, draw a line down, and move an object down.
The programs synthesized are the following:
\begin{Verbatim}[commandchars=\\\{\},fontsize=\tiny, baselinestretch=1]
  (lambda (rotate_cw (draw_line_down (rotate_ccw $0)))) \textcolor{gray}{                         // draw line left}
  (lambda (rotate_cw (move_down (rotate_ccw $0)))) \textcolor{gray}{                              // move object left}
  (lambda (rotate_ccw (draw_line_down (rotate_cw $0)))) \textcolor{gray}{                         // draw line right}
  (lambda (rotate_ccw (move_down (rotate_cw $0)))) \textcolor{gray}{                              // move object right}
  (lambda (rotate_cw (rotate_cw (draw_line_down (rotate_cw (rotate_cw $0)))))) \textcolor{gray}{  // draw line up}
  (lambda (rotate_cw (rotate_cw (move_down (rotate_cw (rotate_cw $0)))))) \textcolor{gray}{       // move object up}
\end{Verbatim}

After running the compression algorithm, the agent creates the following new abstractions:
\begin{Verbatim}[commandchars=\\\{\},fontsize=\tiny, baselinestretch=1]
  (lambda (lambda (rotate_cw ($0 (rotate_ccw $1))))) \textcolor{gray}{                            // apply action left}
  (lambda (lambda (rotate_ccw ($0 (rotate_cw $1))))) \textcolor{gray}{                            // apply action right}
  (lambda (lambda (rotate_cw (rotate_cw ($0 (rotate_cw (rotate_cw $1))))))) \textcolor{gray}{     // apply action up}
\end{Verbatim}

Importantly, the abstractions formed are more general than the original primitives given.
This can help enable systematic generalization on further tasks.

\subsection{Enabling Generalization on ARC Symmetry Tasks} \label{subsec:DC-symmetry}
\begin{figure}
    \centering
    \includegraphics[width=0.7\linewidth]{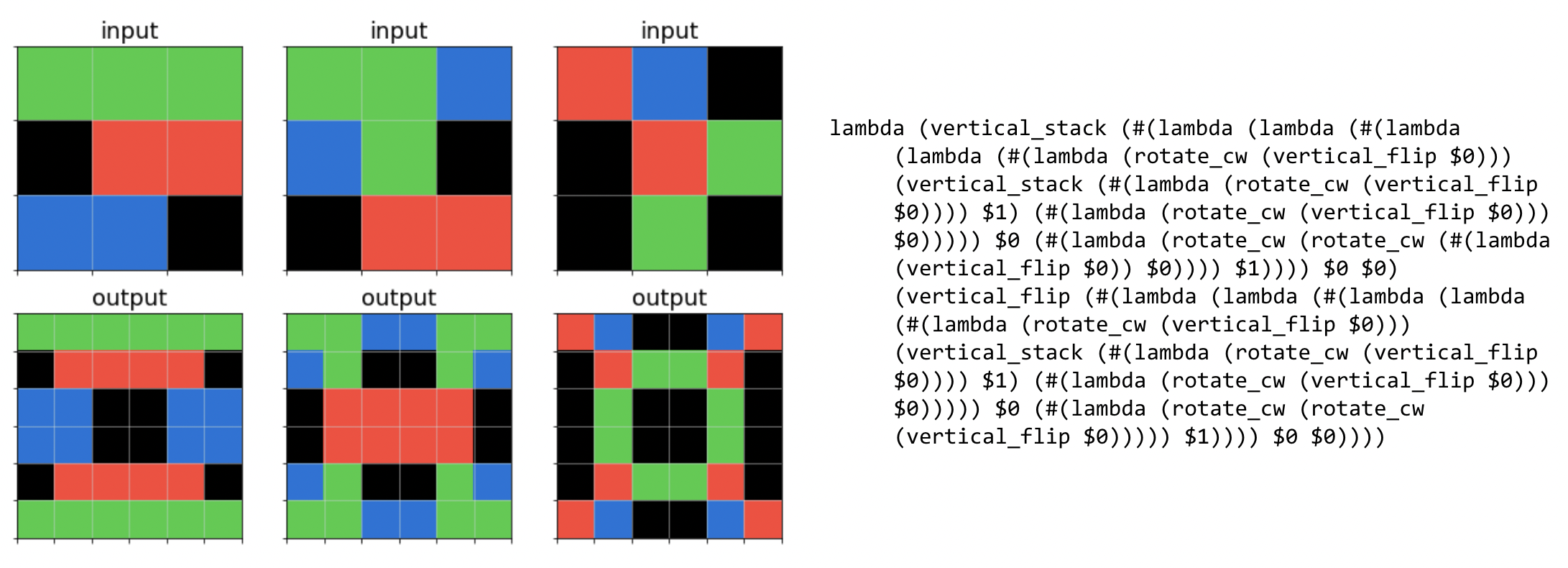}
    \caption[Challenging four-way mirror task]{One of the four-way mirroring tasks and the program discovered that solves it written in terms of the original primitives. The program was discovered only after four iterations of enumeration and compression.}
    \label{fig:symmetry3}
\end{figure}

In a second experiment, we demonstrate how compression-based learning enables developer-aware generalization on ARC. We provide DreamCoder with a set of five grid-manipulation operations: flipping vertically with \verb|vertical_flip|, rotating clockwise with \verb|rotate_cw|, overlaying two grids with \verb|overlay|, stacking two grids vertically with \verb|vertical_stack|, and getting the left half of a grid with \verb|left_half|.
We then train our agent on a subset of 36 ARC tasks involving symmetry over five iterations of enumeration and compression. During each iteration, our agent attempts to solve all 36 tasks by enumerating possible programs for each task. It then runs compression to create new abstractions. During the next iteration, the agent repeats its search equipped with the new abstractions. In this experiment, our agent initially solves 16 tasks. After one iteration, it solves 17 in the same amount of time. After another, it solves 19 tasks, and after the final iteration, it solves 22 tasks. Table \ref{tab:symmetry-abstractions} shows some of the new 
abstractions learned by DreamCoder's compression algorithm such as flipping horizontally, and stacking grids horizontally. The program solutions for the final tasks solved, shown in Figure \ref{fig:symmetry3}, could not be feasibly discovered without the use of abstractions to reduce the search time.

\begin{table}[b]
  \centering
  \begin{tabular}{ll}
    \toprule
    Action & Code \\
    \midrule
    mirror across diagonal & {\scriptsize \verb|#(lambda (rotate_cw (vertical_flip $0)))|} \\
    rotate 180 degrees & {\scriptsize \verb|#(lambda (rotate_cw (rotate_cw $0)))|} \\
    flip horizontally & {\scriptsize \verb|#(lambda (rotate_cw (rotate_cw (vflip $0))))|} \\
    rotate counterclockwise & {\scriptsize \verb|#(lambda (rotate_cw (#(lambda (rotate_cw (rotate_cw $0))) $0)))|} \\
    stack grids horizontally & {\scriptsize \verb|#(lambda (lambda (#(lambda (rotate_cw (vertical_flip $0))) (stack_vertically|} \\
    & {\scriptsize \verb|    (#(lambda (rotate_cw (#(lambda (vertical_flip $0)) $0))) $1)|} \\
    & {\scriptsize \verb|    (#(lambda (rotate_cw (vertical_flip $0))) $0)))))|} \\
    \bottomrule
  \end{tabular}
  \vspace{2mm}
  \caption[Learned symmetry abstractions]{Useful actions learned in the process of solving symmetry tasks. Pound signs represent abstractions. Abstractions may rely on others for construction; e.g. to stack grids horizontally, we reflect each input diagonally, stack vertically, and reflect the vertical stack diagonally.}
  \label{tab:symmetry-abstractions}
\end{table}

\subsection{Discussion} \label{neurips-discussion}
It is useful to compare the learning done in our approach to that done by neural networks. Neural networks can also learn new concepts from training examples, but their internal representation lacks structure which allows them to apply learned concepts compositionally to other tasks. In contrast, functions learned via compression, represented as programs, can naturally be composed and extended to solve harder tasks, while reusing concepts between tasks. This constitutes a learning paradigm which we view as essential to human-like reasoning.


There is a caveat of the approach shown here. Abstraction as shown uses a simple enumerative search. DreamCoder uses a form of neural-guided program synthesis, predicting a distribution over functions to search over, but this guidance is too weak to scale to the complexity of ARC tasks.  In the next section,
we show the type of reasoning required for ARC and design an approach to exhibit this reasoning.

\section{Bidirectional, Neural-guided Program Search}  \label{sec:reasoning}
In Subsection \ref{subsec:bidir} we first motivate and describe our bidirectional, neural-guided search algorithm. Then in Subsection \ref{subsec:bidir-experiments} we present experiments and results using this approach.

\subsection{Algorithm Description} \label{subsec:bidir}
In this section we describe our reasoning approach for ARC. We first give a motivating example of human reasoning on ARC, explain how to approximate it with execution-guided synthesis, then incorporate inverse semantics to create a bidirectional, neural-guided search algorithm. 

\paragraph{Motivating Example}
Solving ARC tasks fundamentally consists of a search for valid solutions. To make this search tractable, our agent needs the ability to reason towards solutions. ARC tasks feature rich visual queues that guide us towards solutions. Without enabling our agent to take full advantage of these queues, the search over possible programs becomes impossibly large. The process of discovering the solution to an ARC task often consists of several discrete steps of reasoning before discovering the solution. How can we design an approach to search that searches for ARC solutions in the same manner as humans?

As a motivating example, let us consider solving task 303 in Figure \ref{fig:task303}. The reasoning steps to come to a solution might look something like this:

\begin{figure}
\centering
 \includegraphics[width=.4\linewidth]{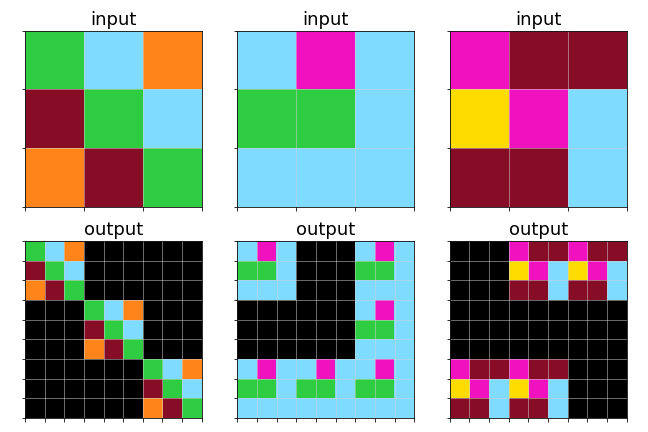}
\caption{Task 303.}
\label{fig:task303}
\end{figure}

\begin{enumerate}
    \item \textbf{Notice} that the output grid consists of copies of the 3x3 input grid, arranged in a certain arrangement among a 9x9 grid.
    \item \textbf{New question:} Where should we place the input grid copies?
    \item \textbf{Notice} that the placements match the arrangement of a different color's pixels for each grid. For example, in the first example, the diagonal of grids in the output matches the green pixels in the input.
    \item \textbf{New question:} What color should we arrange our grid copies along?
    \item \textbf{Solution:} The color matched is the most common color in the grid.
\end{enumerate}

Notice the way discovering a solution involves combining sequential insights and problem reductions. Systematizing a form of reasoning for ARC that emulates this reasoning will be based on a combination of execution-guided program synthesis and inverse semantics.

\begin{figure}
    \centering
    \includegraphics[width=.7\linewidth]{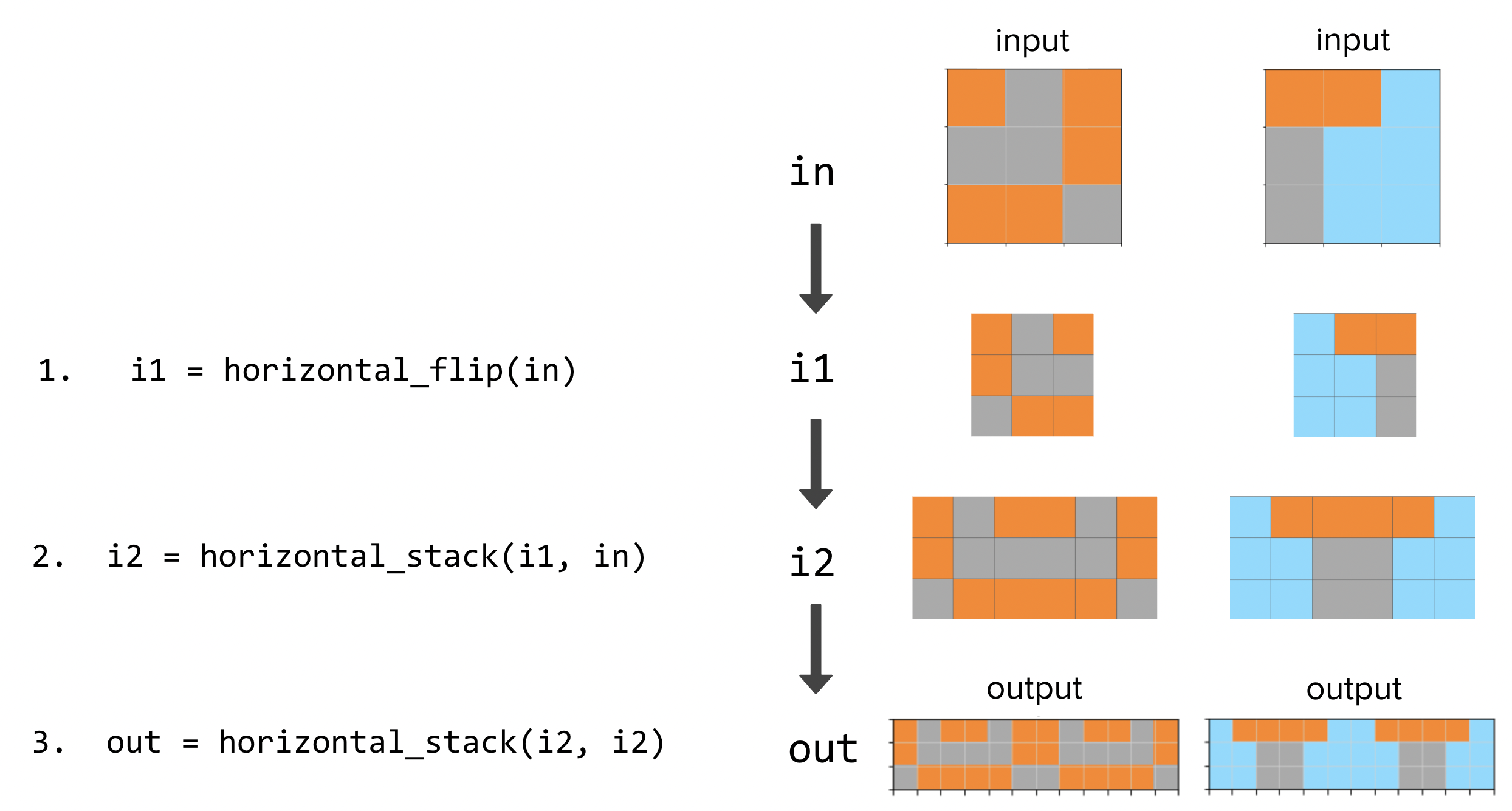}
    \caption[ARC execution-guided walkthrough]{Solving ARC task 138 from the
    evaluation set with execution-guided synthesis. Conditioned on the input
    and output grids, the agent chooses to flip the input horizontally in step
    one. This action is executed to produce intermediate value \texttt{i1}.
    Next, the agent chooses to horizontally stack the intermediate value with
    the input grid, producing another value \texttt{i2}. Last, the agent
    horizontally stacks this value \texttt{i2} with itself, correctly producing the output
    grid for each example and solving the task.}
    \label{fig:repl-example}
\end{figure}

\paragraph{Extending Execution-guided Synthesis}
Execution-guided program synthesis \cite{repl,chen2018execution} is a
form of program synthesis where one executes partial programs to produce
intermediate outputs, which are used to guide the construction of the full program. Intermediate evaluations provide the opportunity for step-by-step reasoning: instead of coming to the answer at once, one can construct it piece by piece. Humans could be said to make use of the same thing: for instance, it much easier to write out the result of a multiplication digit by
digit, instead of conducting the full calculation in one's head.
The form of execution-guided synthesis we apply to ARC is most similar to the `REPL' approach of \cite{repl}.
An example applying the technique to ARC is shown in
Figure \ref{fig:repl-example}. 

Existing execution-based synthesis approaches are limited to bottom-up enumeration: the leaves of the program are constructed (and evaluated) first. In contrast, the steps for solving task 303 involve proposing a
function that is used to produce the output grid, and deducing the inputs required
to correctly produce the output as new intermediate targets before discovering
the complete program.

This form of deductive reasoning involves evaluating function in reverse. It is
best exemplified in the FlashMeta system \cite{flashmeta}, which leverages
the inverse semantics of operators to deduce one or more inputs of a function
given the output target and one or more inputs. We incorporate this type
of reasoning into an extension of execution-guided program synthesis.

\paragraph{Deductive reasoning via inverse semantics}

\begin{figure}
    \centering
    \includegraphics[width=.5\linewidth]{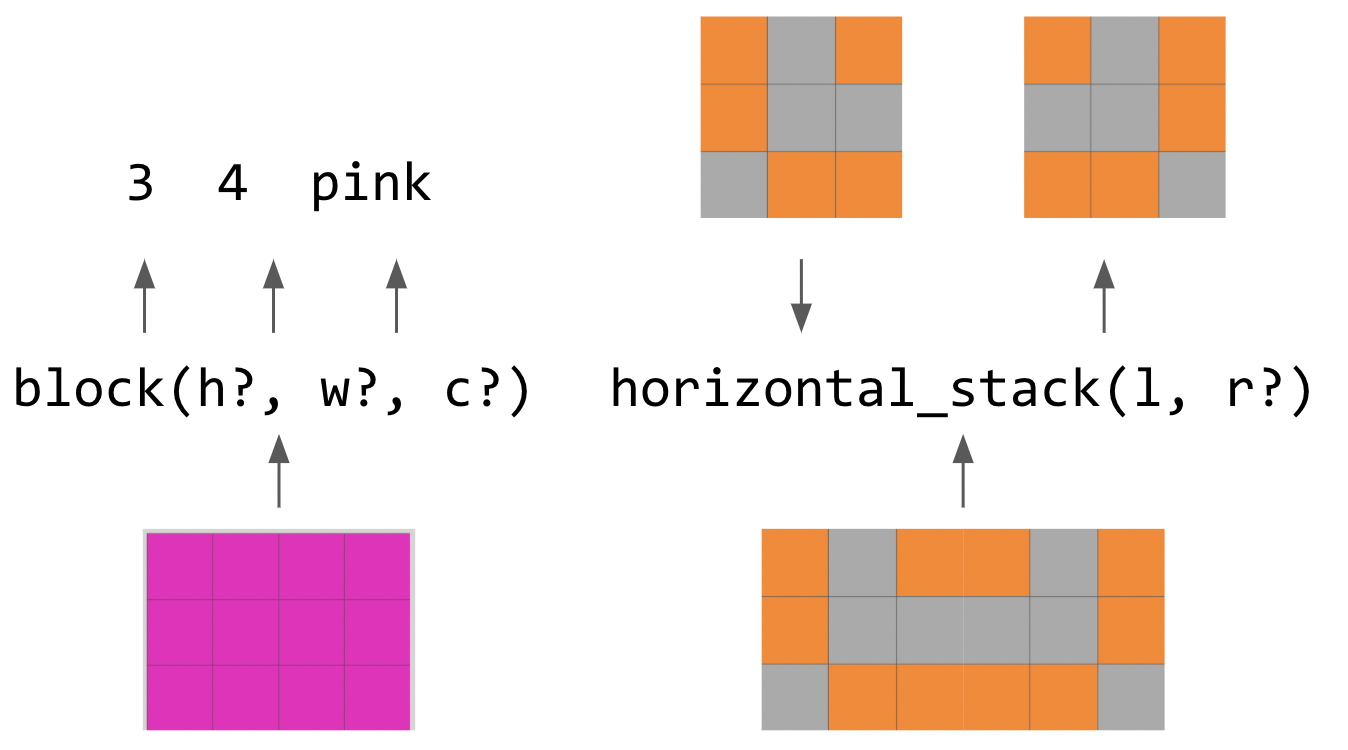}
    \caption[Example inverse and conditional inverse]{\textit{Left:} the function \texttt{block} is directly invertible: given the output, we can deduce the inputs. \textit{Right:} the function \texttt{horizontal$\_$stack} (horizontal stack)
    is conditionally invertible: given the output and one input, we can deduce the other input.}
    \label{fig:inverses}
\end{figure}

For our purposes, we can consider two cases. The simplest case is when the function is invertible. In this case, we
can evaluate the inverse to produce two new targets for the search, as shown in Figure
\ref{fig:inverses}. In the second case, the function is
\textit{conditionally} invertible: given the output and one or more inputs to a function, one can deduce the remaining inputs needed to produce the output via this function. Many functions are conditionally invertible; perhaps the most familiar family is arithmetic operators: if we know $1 + x = 5$, we can deduce that $x = 4$. An example relevant to ARC is shown in Figure \ref{fig:inverses}. Using conditional inverses, it is possible to formalize the reasoning described for task 303.

\paragraph{Bidirectional, Neural-guided Program Search}
To extend execution-guided synthesis to a bidirectional algorithm with inverse and conditional inverse functions, we extend the environment of \cite{repl}, approaching the synthesis task via reinforcement learning.  

The setup takes place in a Markov Decision Process. The current state is a graph of nodes. Each node represents either an input value, the output value, or an intermediate value resulting from the execution of an operation in the forwards or backwards direction. A node is \textit{grounded} if there is a valid program to create that node from the operations applied so far. In general, grounded nodes correspond to those from the forwards, i.e. bottom-up program enumeration, direction of search, while ungrounded nodes correspond to those from the backwards direction, i.e. top-down program enumeration.


An \textit{operation} is a function from the grammar along with a designation of being applied in forwards, inverse, or as a conditional inverse (and if as a conditional inverse, conditioned on which input arguments). There are three types of operations: forward operations, inverse operations, and conditional inverse operations. A forward operation applies the function to a set of grounded inputs to produce a new grounded node. An invertible operation takes an ungrounded output and produces a new ungrounded target node such that grounding the target node will cause the output node to be grounded as well.  A conditionally invertible operation takes an ungrounded output and one or more grounded input nodes, and produces a new ungrounded target node such that grounding the target node will cause the output node to be grounded as well.  All invertible and conditionally invertible operations have a corresponding forward operation.

Solving a given task thus consists of an episode in the MDP. Actions in the MDP correspond to a choice of operation and the choice of arguments for that operation. Each action applies a function in either the forward or backward direction. Intuitively, this executes a bidirectional search to try to connect the grounded nodes on one side with the ungrounded output node on the other. We give reward $R$ for solving the task and a penalty of $-1$ for choosing an action corresponding to an invalid operation.




Like \cite{repl, cmu-repl, chen2018execution}, we train with a combination of supervised training on randomly generated programs fine-tuning with reinforcement learning algorithm \textsc{Reinforce}. To generate random bidirectional programs for supervised training, we first create a random program, and construct an execution trace for it by probabilistically converting inverting function applications from the root. Network architecture is held the same from \cite{repl}, with task-dependent embedding network nodes of the bidirectional graph, a DeepSet network \cite{zaheer2018deep} to encode the graph into a single embedding and choose a function to apply, and a pointer network \cite{vinyals2017pointer} for choosing function arguments.

\subsection{Experiments} \label{subsec:bidir-experiments}
We evaluate our bidirectional algorithm in three settings: solving ARC symmetry tasks, solving arithmetic puzzles from the `24 Game' family, and solving `double-and-add' puzzles. As a baseline, we compare bidirectional synthesis with a forward-only baseline which only allows application of operations in the forwards direction like existing approaches. 


\paragraph{ARC symmetry tasks} As a proof of concept, we evaluate the bidirectional algorithm on a set of 18 ARC symmetry tasks
---a subset of those used in Section \ref{sec:abstraction}. We use a DSL of six operations: stacking two grids horizontally or vertically, rotating clockwise or counterclockwise, and flipping a grid horizontally or vertically. The rotation and flip functions are directly invertible, while the stacking operations are conditionally invertible. We use a convolutional neural network to embed grid example sets.
We train on a set of randomly generated programs evaluated on random input
grids from the ARC training set, and fine-tune with \textsc{Reinforce} before
sampling rollouts for thirty minutes on all tasks at once. The agent is able to
solve 14 of 18 tasks, including one of the ``four-way mirror'' tasks. In this experiment, bidirectional performed equally to the forward-only baseline.



\paragraph{24 Game} Next, we compare the performance of bidirectional search with the forward-only baseline by tasking our agent with solving ``24 Game'' problems. A 24 Game consists of four input numbers, one through nine. To solve the task, one must use each number once in an expression that creates twenty four using $+, -, \times, \div$. For example, given 8, 1, 3, and 2, a
solution is $24 = (2 - 1) \times 3 \times 8$. To solve these tasks bidirectionally, we can use the conditional inverse of each
arithmetic operator in addition to forward arithmetic operations.\footnote{We
relax the rule that each input is used exactly once.}

First we conduct supervised pretraining on all depths at once. These programs may create any
number as a target, not just 24, with the maximum allowed integer 100, and no
negative or nonintegral numbers. We then fine-tune on different depths with
\textsc{Reinforce} for 10,000 epochs of batch size 1000. We measure
performance by percent of episodes solved in the last 1,000 epochs of training.

Results are shown in Table \ref{tab:24-results}. Bidirectional synthesis outperforms
the forward-only baseline across all depths. This supports our thesis, but is suspicious: as we should expect to see identical accuracy for depth one tasks, when only a single action is needed. Accuracy remains fairly high
as depth increases, because depth does not necessarily imply program length: as many as 40\% of depth four tasks remain solvable in fewer than four actions. 

\begin{table}
  \centering
  \begin{tabular}{lllll}
    \toprule
    Depth & 1 & 2 & 3 & 4 \\
    \midrule
    Forward-only & 87.22 {\tiny $\pm 0.64$} & 84.29  {\tiny $\pm 1.6$} & 75.88  {\tiny $\pm 3.6$} & 67.04  {\tiny $\pm 1.0$} \\
    Bidirectional & 95.2  {\tiny $\pm 0.66$} & 92.9  {\tiny $\pm 2.1$} & 87.7  {\tiny $\pm 1.1$} & 85.3  {\tiny $\pm 1.9$} \\ \bottomrule
  \end{tabular}
  \vspace{2mm}
  \caption[Percent of tasks solved for 24 Game]{Percent of tasks solved for 24 Game, measured by percent of episodes solved in the last 1000 epochs of RL fine-tuning. Forward-only denotes only using
    forward operations. Bidirectional includes conditional-inverse operations.
    Average over three runs with stdev shown.}
    \label{tab:24-results}
\end{table}

\begin{figure}
    \centering
    \includegraphics[width=0.7\linewidth]{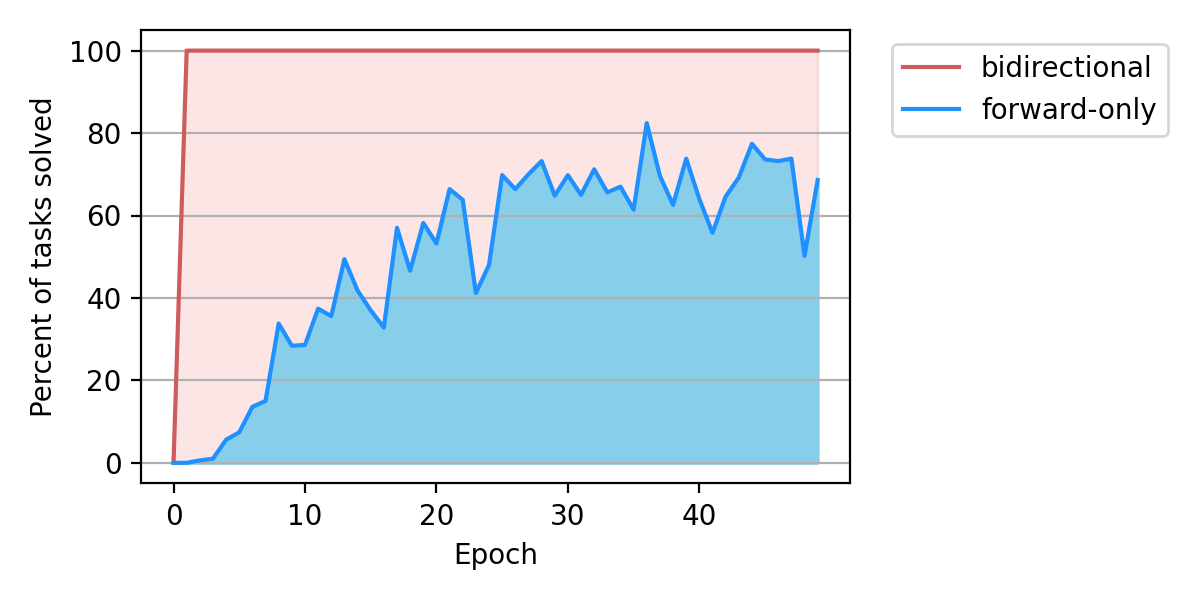}
    \caption[Double-and-add task results]{Percent of tasks solved for bidirectional and forward-only agents trained on the double-and-add task. The bidirectional agent achieves 100\% accuracy after a single epoch of training. After fifty epochs of training, forward-only converges without solving the held-out tasks.}
    \label{fig:binary_figure}
\end{figure}

\paragraph{Double-and-add} Last, we include results on a `double-and-add' task to better show the advantage of bidirectional search. Given a target number, one must reach it starting from the number two by repeatedly adding one or doubling the number. For example, $7 = 1 + 2 * (1 + 2)$. This task, akin to the method for exponentiation by repeated squaring, is much easier solved in a top-down fashion: the choice of adding one or doubling boils down to whether the target is even or odd. Here we have two forward operations, each of which are directly invertible. On a training set of five thousand numbers sampled between one and five million, and a held out set of five hundred numbers, our bidirectional model achieves 100\% evaluation accuracy after a single epoch of supervised training. In contrast, the forward-only model fails to solve the held-out tasks, due to the difficulty ``seeing'' the solution from the source, see Figure \ref{fig:binary_figure}. 


\section{Discussion}\label{sec:discussion}
\paragraph{Related work}
Our work builds off and is inspired by a long line of progress in neural program synthesis \cite{deepcoder, robustfill, cai2017making, nye2019learning, valkov2018houdini}, execution-guided synthesis \cite{repl, chen2018execution, zohar2019automatic}, and deep reinforcement learning for search \cite{silver2017mastering, McAleerASB19}. 

Bidirectional, neural-guided program search is made possible primarily due to the inverse semantics of FlashMeta \cite{flashmeta}. The concept of bidirectional programming, inverse semantics, and program inversion has been present throughout the history of program synthesis \cite{Dijkstra1982, 10.1145/1993316.1993557, Lubin_2020}, but the way in which inverse evaluation is used here is most similar to FlashMeta.

\paragraph{ARC} To date, there are no prominent learning-based approaches to ARC that have proven more successful than the Kaggle-winning brute-force approach \cite{icecuber}. Other Kaggle approaches include genetic programming and cellular automata, but all essentially rely on brute force search over a DSL of operations combined with ARC-specific tricks, without any substantial learning \cite{kaggle}. The few-shot nature and large search space for ARC make it a very challenging benchmark, and progress scaling program synthesis algorithms is likely needed to enable further progress. We hope our progress reported here inspires and enables further progress on ARC.


\paragraph{Future work}
The next step of our work is to combine the two approaches to create a unified approach. This can be done by using the bidirectional search algorithm to solve tasks, then create new operations out of abstractions base on tasks solved. 
To fill out the learning approach, we can consider including the ability to synthesize inverse and conditional inverse operations for newly learned abstractions, perhaps as its own synthesis problem. Our approach remains to be scaled up to a full DSL capable of solving ARC. Incorporating more sophisticated inverse semantics and type-directed search are important components of the full bidirectional approach. 


\bibliography{thesis}
\bibliographystyle{splncs03}
%
%
%







\end{document}